\documentclass[a4paper,fleqn]{cas-dc}

\usepackage[authoryear,longnamesfirst]{natbib}
\usepackage{graphicx}
\usepackage{amsmath}
\usepackage{amssymb}
\usepackage{booktabs}

\usepackage{multirow}
\usepackage{makecell}
\usepackage{color}

\usepackage{algorithm}
\usepackage{algorithmic}

\def\tsc#1{\csdef{#1}{\textsc{\lowercase{#1}}\xspace}}
\tsc{WGM}
\tsc{QE}

\begin{document}
\let\WriteBookmarks\relax
\def\floatpagepagefraction{1}
\def\textpagefraction{.001}

\shorttitle{TPAD: Identifying Effective Trajectory Predictions Under the Guidance of Trajectory AD Model}    

\shortauthors{C. Wang, C. Liang, X. Chen, H. Wang}  

\title [mode = title]{TPAD: Identifying Effective Trajectory Predictions Under the Guidance of Trajectory Anomaly Detection Model}  



%

\author[b]{Chunnan Wang}[style=chinese]
\ead{WangChunnan@hit.edu.cn}





\credit{Credit authorship details}

\author[b]{Chen Liang}[style=chinese]
\ead{1190201818@stu.hit.edu.cn}

\author[b]{Xiang Chen}[style=chinese]
\ead{20s003052@stu.hit.edu.cn}




\author[b]{Hongzhi Wang}[style=chinese,orcid=0000-0002-7521-2871]

\cormark[1]


\ead{wangzh@hit.edu.cn}


\credit{ok}

\affiliation[b]{organization={Harbin Institute of Technology},
            city={Harbin},
            country={China}}

\cortext[1]{Hongzhi Wang}



\begin{abstract}
Trajectory Prediction (TP) is an important research topic in computer vision and robotics fields. Recently, many stochastic TP models have been proposed to deal with this problem and have achieved better performance than the traditional models with deterministic trajectory outputs. However, these stochastic models can generate a number of future trajectories with different qualities. They are lack of self-evaluation ability, that is, to examine the rationality of their prediction results, thus failing to guide users to identify high-quality ones from their candidate results. This hinders them from playing their best in real applications. In this paper, we make up for this defect and propose TPAD, a novel TP evaluation method based on the trajectory Anomaly Detection (AD) technique. In TPAD, we firstly combine the Automated Machine Learning (AutoML) technique and the experience in the AD and TP field to automatically design an effective trajectory AD model. Then, we utilize the learned trajectory AD model to examine the rationality of the predicted trajectories, and screen out good TP results for users. Extensive experimental results demonstrate that TPAD can effectively identify near-optimal prediction results, improving stochastic TP models' practical application effect.
\end{abstract}



\maketitle

\section{Introduction}
\label{sec:intro}

Human Trajectory Prediction (TP) which aims to predict the movement of pedestrians has become a research hotpot recently. Many effective TP solutions have been proposed to resolve this task, and they have been widely applied to many applications, such as autonomous driving~\cite{DBLP:journals/ral/LuoCBHLM18} and surveillance systems~\cite{DBLP:journals/pami/PflugfelderB10}.

Despite the success of these existing TP works, we notice that the state-of-the-arts TP models are generally stochastic. They fail to maintain a high performance in the real applications due to lack of effective distinguish method. Specifically, the TP models proposed in recently years generally add the random Gaussian noise to their predictions~\cite{DBLP:journals/pr/ZamboniKGNC22,DBLP:journals/pr/BarataNLM21,DBLP:conf/cvpr/PangZ0W21}, or model the pedestrian trajectories as a bi-variate Gaussian distribution~\cite{DBLP:journals/pr/HuangWPSY21,DBLP:journals/pr/PeiQZMY19} and estimate the parameters of the distribution. These stochastic factors allow TP models to generate diverse future trajectories for pedestrians. And the experimental results have demonstrated that the best-performed sample of these stochastic TP models can achieve high prediction accuracy, outperforming the deterministic trajectories predicted by the traditional TP models~\cite{DBLP:conf/cvpr/XuPG18,DBLP:conf/icra/VemulaMO18,2018TrafficPredict,DBLP:conf/cvpr/Liang0NH019}.

However, the existing stochastic TP models fail to identify the best-performed samples from number of when the real future trajectories are unknown. They lack effective distinguish method to cope with practical application scenarios, and this reduces their practicability. Specifically, we notice that different trajectory prediction results w.r.t. a certain input, that are generated by a stochastic TP model, differ a lot in terms of accuracy. In the real applications, the existing stochastic TP model can only randomly select some of the prediction results as the final outputs. And this makes their performance instable. 

In order to make up for this defect, in this paper, we propose TPAD, a novel TP evaluation method based on the trajectory \underline{A}nomaly \underline{D}etection (AD) technique, to help the stochastic TP models to identify high-performance trajectory prediction results.

In TPAD, we introduce AD techniques, which can learn effective laws from known normal data and identify abnormal data~\cite{DBLP:journals/pr/DoshiY21,DBLP:journals/pr/ChangTXLZSY22,DBLP:journals/pr/ChenYCXJ21}. We aim to construct a trajectory AD model to learn effective laws from known trajectory data, and thus effectively quantify the rationality of the predicted trajectories during the testing and using phase. This auxiliary model can help us reasonably analyze the effectiveness of the trajectory prediction results without knowing real future trajectories. It can guide the stochastic TP models to output high-quality results in practical use, and thus achieve good application effect.

However, current AD techniques are generally designed for videos and images~\cite{DBLP:journals/pr/ChangTXLZSY22,DBLP:journals/pr/HaoLWWG22,DBLP:journals/pr/ZavrtanikKS21}, and hardly involved in the field of pedestrian trajectory. There is a lack of effective trajectory AD model to support our idea. In order to solve this problem, we utilize the design experience in AD and TP field, and combine \underline{Auto}mated \underline{M}achine \underline{L}earning (AutoML) technique~\cite{DBLP:journals/pr/Tian0XJY21, DBLP:conf/ijcai/GaoYZ0H20, DBLP:journals/pr/ZhangZZXWH21} to independently design AD model for TP area. Specifically, the design experience of the existing AD and TP models provides us with a large number of components for designing trajectory AD model. And the AutoML technique makes it possible to automatically search for effective trajectory AD models. The combination of three fields of techniques enables us to easily obtain powerful trajectory AD model.  

In TPAD, we let the learned trajectory AD model participate in the  decision-making phase of the stochastic TP model. We utilize trajectory AD model to calculate the abnormality scores of various trajectory prediction results, and select the results with low abnormality scores as the final outputs. In this way, the stochastic TP model can maintain a high performance under the guidance of abnormality analysis, showing good applicability in real trajectory scenes.

The experimental results show that the trajectory AD model we designed is effective, and our TPAD method can effectively identify near-optimal prediction results of the stochastic TP models. With the help of TPAD, the existing stochastic TP models can achieve much better application effect, playing out their high performance.

Our major contributions are concluded as follows.

\begin{itemize}
\item[1.] \textit{Innovation}: We innovatively introduce anomaly detection technique to TP area, and present a novel method to guide the reasonable use of stochastic TP models.

\item[2.] \textit{Multi-Technique Integration}: We flexibly use and combine the experience in the fields of AutoML, AD and TP, and build an effective AD model for pedestrian trajectory. The idea of integrating existing techniques to efficienntly solve a new problem is interesting and meaningful.

\item[3.] \textit{Effectiveness}: Extensive experiments show that our designed TP evaluation method can effectively identify high-performance prediction results, which demonstrate the effectiveness of TPAD.
\end{itemize}


\begin{figure*}
\begin{center}
\includegraphics[width=1.0\textwidth, height=0.5\textwidth]{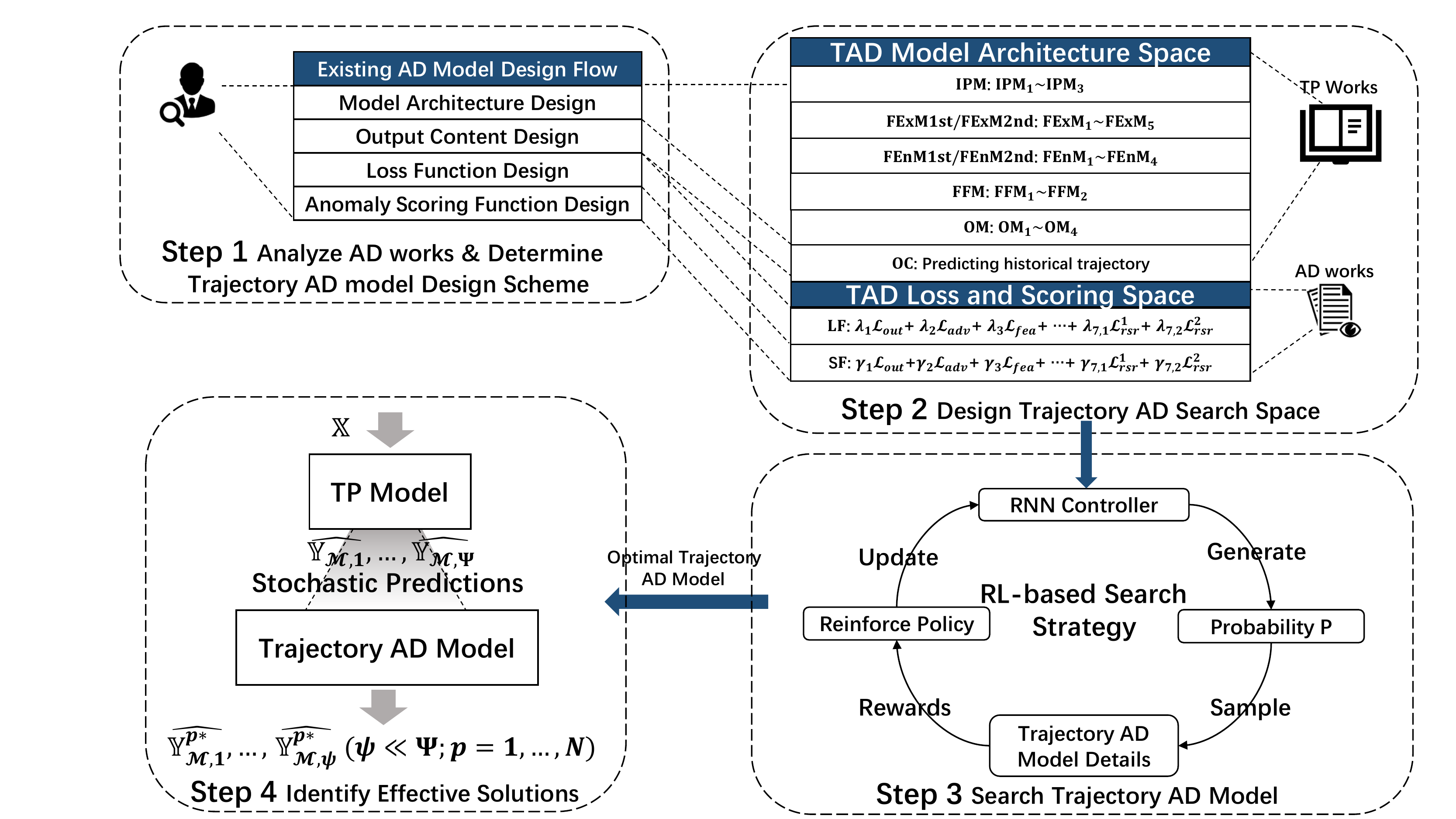}
\end{center}
\caption{The workflow of the TPAD algorithm. $\widehat{\mathbb{Y}_{\mathcal{M},i}^{p*}}\in \mathbb{R}^{t_{pred}\times 2}\ (i=1,\ldots,\psi; p=1,\ldots,N)$ in Step 4 denotes the $i$-th best trajectory prediction result of the $p$-th pedestrian in the given trajectory scene.}
\label{fig3}
\end{figure*}

\section{Related Works}
In this section, we introduce the existing TP models and the existing methods of identifying high-quality TP results.

\textbf{TP Models.}  Human TP which learns to accurately predict pedestrians’ future trajectories given their historical trajectory data~\cite{mohamed2020social}, plays an important role in autonomous driving~\cite{DBLP:journals/ral/LuoCBHLM18} and surveillance systems~\cite{DBLP:conf/icra/LuberSTA10}. Many TP models have been developed to address this task, and among them deep neural network based solutions~\cite{amirian2019social,kosaraju2019social} are the most popular. In recent years, researchers have proposed many effective neural networks, for example, recurrent neural network based generative adversarial model, spatial-temporal graph attention network and graph convolutional neural network, to model motion pattern of each pedestrian~\cite{alahi2016social}, pedestrians’ interactions~\cite{huang2019stgat} and motion tendency~\cite{DBLP:conf/cvpr/Shi0LZZN021} et al. These TP models utilize the learned features to reasonably predict future trajectories, and have made promising progress in real applications.

\textbf{Identification of Effective TP Results.} According to the prediction method, existing TP models can be classified into two categories: deterministic ones~\cite{DBLP:conf/cvpr/XuPG18,DBLP:conf/icra/VemulaMO18,2018TrafficPredict,DBLP:conf/cvpr/Liang0NH019} and stochastic ones~\cite{DBLP:journals/pr/BarataNLM21,mohamed2020social}. The former ones output the deterministic future trajectory for each pedestrian. This kind of TP model can directly provides the optimal prediction result without any assistance.

As for the stochastic TP models, they can generate diverse future trajectories with different qualities for each pedestrian. This kind of TP model (1) adds the random Gaussian noise to their predictions~\cite{DBLP:conf/eccv/YuMRZY20,DBLP:journals/pr/ZamboniKGNC22,DBLP:conf/cvpr/PangZ0W21}, or (2) models the pedestrian trajectories as a bi-variate Gaussian distribution~\cite{mohamed2020social,shi2021sgcn} and estimate the parameters of the distribution. These stochastic factors enable TP models to generate more accurate predictions results, but also bring many poor ones.

The existing stochastic TP models fail to identify high-quality prediction results when the ground-truth are unknown~\cite{DBLP:journals/pr/BarataNLM21,DBLP:conf/cvpr/PangZ0W21}. They can only randomly select some prediction results for practical use, and this makes their performance unstable in real applications. In this paper, we try to solve this problem. We aim to design an effective method to identify high-quality prediction results without the help of the ground-truth outputs, improving the practicability of the stochastic TP models.


\begin{table*}[t]
	\caption{Design process of the existing AD models. Solutions of the each stage are extracted from 5 state-of-the-arts AD models, including P-Net ~\cite{DBLP:conf/eccv/ZhouXYCLLGLG20} ($\mathcal{A}1$), MNAD~\cite{park2020learning} ($\mathcal{A}2$), MPED-RNN~\cite{DBLP:conf/cvpr/MoraisL0SMV19} ($\mathcal{A}3$), GEPC~\cite{markovitz2020graph} ($\mathcal{A}4$) and RSRAE~\cite{DBLP:conf/iclr/LaiZL20} ($\mathcal{A}5$).}
	\newcommand{\tabincell}[2]{\begin{tabular}{@{}#1@{}}#2\end{tabular}}
	\centering
	\resizebox{1.0\textwidth}{!}{
	\begin{tabular}{m{2.5cm}m{3.5cm}m{5.2cm}m{8.6cm}}
    \toprule
		\textbf{Stage} & \textbf{Function Description} & \textbf{Involved Operations} &  \textbf{Solutions Provided by Existing Works} \\
    \midrule
		\multirow{2}{\linewidth}{\textbf{Stage 1: Model Architecture Design}}
		& \multirow{2}{\linewidth}{Design the architecture of the AD model.}
		& \makecell[l]{Feature Extraction Module ($\mathbf{FExM}$)\\ $\mathbb{H}=\mathbf{FExM}(\mathbb{I})$}&  \makecell[l]{ $\mathbf{FExM}_1$: Subsampled CNN encoder ($\mathcal{A}_1,\mathcal{A}_2$)
			\\
			$\mathbf{FExM}_2$: RNN encoder ($\mathcal{A}_3$)
			\\
			$\mathbf{FExM}_3$: GNN-based encoder ($\mathcal{A}_4$)
			\\
			$\mathbf{FExM}_4$: Linear encoder ($\mathcal{A}_5$)
		}  \\     
		\cline{3-4} 	
		&&\makecell[l]{Feature Enhancement Module ($\mathbf{FEnM}$)\\ $\mathbb{H}^{\prime}=\mathbf{FEnM}(\mathbb{H})$}&  \makecell[l]{ $\mathbf{FEnM}_1$: Identity ($\mathcal{A}_1,\mathcal{A}_3,\mathcal{A}_4$)
			\\
			$\mathbf{FEnM}_2$: Memory module ($\mathcal{A}_2$)
			\\
			$\mathbf{FEnM}_3$: Linear transformation module ($\mathcal{A}_5$)
		} \\
		\cline{3-4} 	
		&&\makecell[l]{Output Module ($\mathbf{OM}$)\\ $\mathbb{O}'=\mathbf{OM}(\mathbb{H}^{\prime})$}&  \makecell[l]{ $\mathbf{OM}_1$: Up-sampled CNN decoder ($\mathcal{A}_1,\mathcal{A}_2$)
			\\
			$\mathbf{OM}_2$: RNN decoder ($\mathcal{A}_3$)
			\\
			$\mathbf{OM}_3$: GNN-based decoder with a clustering layer ($\mathcal{A}_4$)
			\\
			$\mathbf{OM}_4$: Linear decoder ($\mathcal{A}_5$)
		} \\

		\midrule
		\textbf{Stage 2: Output Content Design}
		 & Determine the output content of AD model.   &\makecell[l]{ Output Content ($\mathbf{OC}$)
			}
		&  \makecell[l]{
		$\mathbf{OC}_1$: Predicting related content of the input ($\mathcal{A}_3$)
			\\
			$\mathbf{OC}_2$: Reconstructing the input ($\mathcal{A}_1,\mathcal{A}_2,\mathcal{A}_3,\mathcal{A}_4,\mathcal{A}_5$) 
		}
		\\     \midrule

		\textbf{Stage 3: Loss Function Design}                & Design an effective loss function for AD model training. &  \makecell[l]{Loss Function ($\mathbf{LF}$)\\$\mathcal{L}=\mathbf{LF}(\mathbb{O},{\mathbb{O}^{\prime}}, \mathbb{H}^{\prime})$} &\makecell[l]{$\mathbf{LF}_1$: Output error + Adversial loss + Feature difference ($\mathcal{A}_1$)
			\\
			\makecell[l]{$\mathbf{LF}_2$: Output error + MNAD compactness loss + MNAD\\ \ \ \ \ \ \ \ separateness loss ($\mathcal{A}_2$)}
			\\
			$\mathbf{LF}_3$: Output error ($\mathcal{A}_3$)
			\\
			$\mathbf{LF}_4$: Output error + GEPC Clustering loss ($\mathcal{A}_4$)
			\\
			$\mathbf{LF}_5$: Output error + RSR loss ($\mathcal{A}_5$)
		} \\     \midrule

		\multirow{2}{\linewidth}{\textbf{Stage 4: Anomaly Scoring Function Design}} & \multirow{2}{\linewidth}{Design an effective anomaly scoring function to examin the rationality of the given input.}   & \makecell[l]{Scoring Function ($\mathbf{SF}$) \\$\mathcal{F}=\mathbf{SF}(\mathbb{O},{\mathbb{O}^{\prime}}, \mathbb{H}^{\prime})$}
		&\makecell[l]{$\mathbf{SF}_1$: Output error + Feature difference ($\mathcal{A}_1$)
		\\
		$\mathbf{SF}_2$: Output error + MNAD compactness loss ($\mathcal{A}_2$)
		\\
		$\mathbf{SF}_3$: Output error ($\mathcal{A}_3$)
		\\
		$\mathbf{SF}_4$: Output error + GEPC Clustering loss ($\mathcal{A}_4$)
		\\
		$\mathbf{SF}_5$: Output error + RSR loss ($\mathcal{A}_5$)
		}
		\\     
	\bottomrule
	\end{tabular}
	}
	\label{table1}
\end{table*}

\section{Our Approach}
In this section, we design the TPAD method to guide the rational use of the stochastic TP models. Firstly, we analyze the existing AD methods, determining the design flow of the trajectory AD models. (Section~\ref{section:3.1}). Secondly, we design an effective search space to suppport the automatic design of the trajectory AD model (Section~\ref{section:3.2}). Thirdly, we find a suitable AutoML algorithm to search for the powerful trajectory AD models (Section~\ref{section:3.3}). Finally, we utilize the learned trajectory AD model to help the existing TP models identify effective solutions (Section~\ref{section:3.4}). Figure~\ref{fig3} gives the overall workflow of our TPAD method.

\subsection{Notations and Design Idea}\label{section:3.0}

\textbf{Notations on TP.} Assume there are $N$ pedestrians involved in a scene, represented as $p_1, p_2, \ldots, p_N$. The position of pedestrian $p_i$ at time-step $t$ is denoted as $p_i^t=\left(x_i^t,y_i^t\right)$, and the set of observed history positions of all pedestrians over a time period $t_{\mathrm{obs}}$ is denoted as $\mathbb{X}=\left\{p_i^{1:t_{\mathrm{obs}}}\middle| i=1,\ldots,N\right\}\in \mathbb{R}^{N \times t_{obs}\times 2}$. A stochastic TP model $\mathcal{M}$ can predict the upcoming trajectories of all pedestrians over a future time horizon $t_{\mathrm{pred}}$, which is denoted by $\mathbb{Y}=\left\{p_i^{t_{\mathrm{obs}}+1:t_{\mathrm{pred}}}\middle| i=1,\ldots,N\right\}\in \mathbb{R}^{N \times t_{pred}\times 2}$, according to $\mathbb{X}$. We use $\widehat{\mathbb{Y}_{\mathcal{M},i}}=\mathcal{M}\left(\mathbb{X}\right)$ to represent a prediction result provided by the stochastic model $\mathcal{M}$, and compare $ \widehat{\mathbb{Y}_{\mathcal{M},i}}\ (i=1,\ldots,\Psi)$, where $\Psi$ denotes the stochastic sample number, with $\mathbb{Y}$ using a certain loss function, so as to examine the effectiveness of model $ \mathcal{M}$. 

\textbf{Design Idea of TPAD.} In TPAD, we aim to construct an effective trajectory AD model $\mathcal{F}$ to differentiate between prediction results with different precisions. We hope $\mathcal{F}$ to give the lowest anomaly score to the real future trajectories, while higher anomaly scores to the prediction results with lower precisions:
\begin{equation}
\begin{split}
\mathbf{avg}[\mathcal{F}(&\mathbb{Y},\mathbb{X})] < \mathbf{avg}[\mathcal{F}(\widehat{\mathbb{Y}_{\mathcal{M},i}},\mathbb{X})]  \\
&s.t.\ \Vert \mathbb{Y},\widehat{\mathbb{Y}_{\mathcal{M},i}} \Vert_2 > 0
\end{split}
\end{equation}
where $\mathcal{F}(\cdot,\mathbb{X})\in \mathbb{R}^{N}$ denotes the anomaly score of the prediction result on each pedestrian, and $\mathbf{avg}[\mathcal{F}(\cdot,\mathbb{X})]\in \mathbb{R}$ is the average anomaly score on all pedestrians. In this way, high-quality prediction results can be identified under the guidance of anomaly scores during the test and use phase.

Based on the above idea, we apply Area Under ROC Curve ($\mathbf{AUC}$) to examine the effectiveness of a trajectory AD model, and define the design target of TPAD as follows. 

\textbf{Design Target.} Given a TP dataset $\mathbb{D}=\{\mathbb{D}_{train}, \mathbb{D}_{val}\}$ and a trajectory AD search space $\mathbb{S}$, TPAD aims to find an optimal trajectory AD model $\mathcal{F}^\ast \in \mathbb{S}$ with the highest $\mathbf{AUC}$ score on the validation set:
\begin{equation}
	\begin{aligned}
		\mathcal{F}^\ast&=\max_{\mathcal{F}\in\mathbb{S}} {\mathbf{AUC}_{\mathbb{D}_\mathrm{val}\bigcup \mathbb{D}_\mathrm{val}^{-}}} (\mathbf{W}_\mathcal{F}^\ast,\mathcal{F})\\
		&= \max_{\mathcal{F}\in\mathbb{S}} \mathbf{P}(\mathbf{avg}[\mathcal{F}(\mathbb{Y},\mathbb{X};\mathbf{W}_\mathcal{F}^\ast)] <\mathbf{avg}[\mathcal{F}(\mathbb{Y}^{-},\mathbb{X};\mathbf{W}_\mathcal{F}^\ast)] )\\
		&\mathrm{s.t.}\ \mathbf{W}_\mathcal{F}^\ast= \mathop{\arg\min}_{\mathbf{W}}{\mathcal{L}_{\mathbb{D}_\mathrm{train}}}(\mathbf{W},\mathcal{F}) \\
	\end{aligned}
\end{equation}
where $\mathbb{D}_{train}$ and $\mathbb{D}_{val}=\{(\mathbb{Y}, \mathbb{X})\}$ are positive sample set with normal pedestrian trajectories, and $\mathbb{D}_{val}^{-}=\{(\mathbb{Y}^{-},\mathbb{X})\}$ is a negative sample set, constructed by adding noises to the real future trajectories in $\mathbb{D}_{val}$. 

In TPAD, we will design an effective search space $\mathbb{S}$ and find a suitable AutoML method, so as to automatically design $\mathcal{F}^\ast$. Then, we will design an effective method based on $\mathcal{F}^\ast$ to guide the rational use of stochastic TP models.

\subsection{Design Method of Trajectory AD Model}\label{section:3.1}
In this part, we analyze the existing AD models and aim to find a suitable method to design our trajectory AD model.

\textbf{Design Process of Existing AD Models.} We analyze the design process of the existing AD models, and sum up the following 4 stages of the AD model design. We extract effective solutions of each stage from state-of-the-art AD models (as is shown in Table~\ref{table1}), and try to find suitable design modules for trajectory AD model.

\begin{table*}[t]
	\caption{Detailed introduction of the loss functions mentioned in Table~\ref{table1}.}
	\newcommand{\tabincell}[2]{\begin{tabular}{@{}#1@{}}#2\end{tabular}}
	\centering
	\resizebox{1.0\textwidth}{!}{
	\begin{tabular}{m{1.2cm}m{3.4cm}m{7.5cm}m{9.5cm}}
    \toprule
		\textbf{Symbols} & \textbf{Loss Function} & \textbf{Function in Training Phase} &  \textbf{Function Details} \\
    \midrule
		$\mathcal{L}_{out}$ & Output error & Minimize the distance between the model output and the ground truth. & \makecell[l]{$\mathcal{L}_{out}=\Vert \mathbb{O}-\mathbb{O}'\Vert_2$} 
		\\   
		$\mathcal{L}_{adv}$ & Adversial loss &  Guide model to synthesize more realistic outputs with a discriminator ($\mathbf{Dis}$). & \makecell[l]{$\mathcal{L}_{adv}=\mathbb{E}[log(\mathbf{Dis}(\mathbb{I},\mathbb{O}'))]$}
		\\   
		$\mathcal{L}_{fea}$ & Feature difference & Minimize the distance between prediction's feature and the ground-truth output's feature (two features are calculated by $\mathbf{Dis}$'s feature extractor, denoted by $\mathbf{Dis_{fea}}$). &  \makecell[l]{$\mathcal{L}_{fea}=\Vert \mathbf{Dis_{fea}}(\mathbb{I},\mathbb{O})-\mathbf{Dis_{fea}}(\mathbb{I}, \mathbb{O}^\prime)\Vert_2$} 		
		\\   
		$\mathcal{L}_{com}$ & MNAD compactness loss & Encourage the query features extracted from the normal data to be close to the nearest item in the memory module, reducing intra-class variations. & \makecell[l]{$\mathcal{L}_{com}=\sum_{k=1}^{K} \Vert \mathbf{q}^{k}-\mathbf{p}_{p} \Vert_2$ \\ $\mathbf{q}^{k}$: a query feature of $\mathbf{I}$ extracted by $\mathbf{FEnM}_2$; \\$\mathbf{p}_{p}:$ the nearest item for $\mathbf{q}^{k}$ in memory module} 
		\\   
		$\mathcal{L}_{sep}$ & MNAD separateness loss & Make the items in the memory far enough apart from each other to consider various patterns of normal data. & \makecell[l]{$\mathcal{L}_{sep}=\sum_{k=1}^{K} [\Vert \mathbf{q}^{k}-\mathbf{p}_{p} \Vert_2-\Vert \mathbf{q}^{k}-\mathbf{p}_{n} \Vert_2]$ \\ $\mathbf{p}^{n}$: the second nearest item for $\mathbf{q}^{k}$ in memory module} 
		\\   
		$\mathcal{L}_{clu}$ & GEPC Clustering loss & Strengthen cluster assignments of model on normal data by normalizing and pushing each value closer to a value of either 0 or 1. & \makecell[l]{$\mathcal{L}_{clu}=\sum_{c=1}^{C} \mathbf{b}_{c}log\frac{\mathbf{b}_{c}}{\mathbf{d}_{c}}$ \\ $\mathbf{b_{c}}$: probability that $\mathbb{I}$ is assigned to the $c$-th cluster,  calculated by $\mathbf{OM}_3$; \\$\mathbf{d}_{c}$: the target hard-assignment following~\cite{DBLP:conf/icml/XieGF16}} 
		\\    
		$\mathcal{L}_{rsr}^{1}, \mathcal{L}_{rsr}^{2}$ & RSR loss & Learn the hidden linear structure of the embedded normal points, mapping them near their original locations (anomalies will lie away from their locations). & \makecell[l]{$\mathcal{L}_{rsr}=\mu_{1}\mathcal{L}_{rsr}^{1}+\mu_{2}\mathcal{L}_{rsr}^{2}$\\$\ \ \ \ \ =\mu_{1}\Vert \mathbb{H}-\mathbf{A}^{\mathrm{T}}\mathbf{A}\mathbb{H}\Vert_2^{2}+\mu_{2}\Vert \mathbf{A}\mathbf{A}^{\mathrm{T}}-\mathbf{I}\Vert_F^{2}$\\$\mathbf{A}$:  a linear transformation matrix defined in $\mathbf{FEnM}_3$;\\ $\mathbf{I}$: an identity matrix}
		\\     
	\bottomrule
	\end{tabular}
	}
	\label{table2}
\end{table*}

\textbf{Stage 1: Model Architecture Design.} The existing AD model architectures are composed of three parts: Feature Extraction Module ($\mathbf{FExM}$), Feature Enhancement Module ($\mathbf{FEnM}$) and Output Module ($\mathbf{OM}$). The structure type of these modules is determined by the AD object. 

For example, MNAD~\cite{park2020learning} ($\mathcal{A}2$) is designed for AD tasks on images, and applies CNN networks to construct the AD model. GEPC~\cite{markovitz2020graph} ($\mathcal{A}4$) uses the GNN architectures since its AD object is the human pose graph.

\textbf{Stage 2: Output Content Design.} There are two types of AD model outputs: input reconstruction and relevant content prediction. The former one can guide the AD model to learn normal samples' effective features\footnote{These features can either be extracted from the input sample or be used to reconstruct the input.} by minimizing the reconstruction error. This kind of output is suitable for AD tasks where anomalous have completely different meanings or behaviors. In these tasks, anomalous will be poorly reconstructed and identified due to different features.

The latter one guides the AD model to learn the mapping rule from the normal sample to the known content related to it, by minimizing the prediction error. For example, MPED-RNN~\cite{DBLP:conf/cvpr/MoraisL0SMV19} ($\mathcal{A}3$) analyzes the pose sequence and takes the future or past poses as the output. This kind of output is suitable for AD tasks where the difference between anomalous and normal ones may be small, but can influence the prediction precision. In these tasks, anomalous will be poorly predicted and identified due to unsuitable mapping rules.

\textbf{Stage 3: Loss Function Design.} Minimizing the output error is not the only way of training AD models. Researchers have proposed many other loss functions to guide the AD model training (as is shown in Table~\ref{table2}). 

For example, GEPC~\cite{markovitz2020graph} ($\mathcal{A}4$) presents the clustering loss $\mathcal{L}_{clu}$ to learn the hard cluster assignment of the normal outputs. MNAD~\cite{park2020learning} ($\mathcal{A}2$) designs the $\mathcal{L}_{com}$ and $\mathcal{L}_{sep}$ to extract various normal patterns as the basis for AD. The addition of these loss functions enables the AD models to gain more AD criterions after training and thus achieve better performance.

In addition, we notice that AD models may need to add some auxiliary modules for using these loss functions. For example, $\mathcal{L}_{clu}$ relies on the memory module design by $\mathcal{A}2$, and $\mathcal{L}_{clu}$ relies on the clustering layer designed by $\mathcal{A}4$. These auxiliary modules will be added to the AD model with the use of their corresponding loss functions.

\textbf{Stage 4: Anomaly Scoring Function Design.} We can see from Table~\ref{table1} that the existing AD models generally use some components of the loss function to construct the anomaly scoring function. They give normal samples with small loss values  and very small anomaly scores, and consider samples with high anomaly scores as the anomalous.

\textbf{Design Process of Trajectory AD model.} In TPAD, we can follow the above four stages to design the trajectory AD model. But need to make some adjustments according to characteristics of the application scenes of the trajectory AD model, so as to obtain a more effective design scheme. 

We select suitable solutions for each stage from the existing AD works and TP works, and thus obtain the following \underline{t}rajectory \underline{AD} (TAD) model design scheme.

\begin{table*}[t]
	\caption{Design process of the trajectory AD model. Solutions of the each stage are extracted from two parts: (1) 5 state-of-the-arts AD models analyzed in Table~\ref{table1}; (2) 5 state-of-the-art TP models, including SGCN~\cite{DBLP:conf/cvpr/Shi0LZZN021} ($\mathcal{M}_1$), LB-EBM~\cite{DBLP:conf/cvpr/PangZ0W21} ($\mathcal{M}_2$), Social-STGCNN~\cite{mohamed2020social} ($\mathcal{M}_3$), Social Ways~\cite{amirian2019social} ($\mathcal{M}_4$), STGAT~\cite{huang2019stgat} ($\mathcal{M}_5$). }
	\newcommand{\tabincell}[2]{\begin{tabular}{@{}#1@{}}#2\end{tabular}}
	\centering
	\resizebox{1.0\textwidth}{!}{
	\begin{tabular}{m{2.5cm}m{3.5cm}m{5.5cm}m{8.6cm}}
    \toprule
		\textbf{Stage} & \textbf{Function Description} & \textbf{Involved Operations} &  \textbf{Solutions or Options used for Search Space Construction} \\
    \midrule
		\multirow{5}{\linewidth}{\textbf{Stage 1: TAD Model Architecture Design}}
		& \multirow{5}{\linewidth}{Design the architecture of the trajectory AD model.}
		&\makecell[l]{Input Processing Method (IPM)\\ $\ \ \ \ \mathbb{I}^{\prime}=\mathbf{IPM}(\mathbb{I})$\\
		$\ \ \ \ \mathbb{I}^{\prime} \in \mathbb{R}^{N\times t_{pred}\times 2}$} &   \makecell[l]{ $\mathbf{IPM}_1$: Real Position ($\mathcal{M}_2,\mathcal{M}_3$)
			\\
			$\mathbf{IPM}_2$: Relative Position ($\mathcal{M}_1,\mathcal{M}_5$)
			\\
			$\mathbf{IPM}_3$: Real + Relative Position ($\mathcal{M}_4$)
		}   \\     
		\cline{3-4} 	
		&&\makecell[l]{\\Feature Extraction Method (FExM)
			\\$\ \ \ \ \mathbb{H}_{1st} =\mathbf{FExM}{\mathbf{1st}}(\mathbb{I}^\prime)$
			\\$\ \ \ \ \mathbb{H}_{2nd} =\mathbf{FExM}{\mathbf{2nd}}(\mathbb{I}^\prime)$
			\\$\ \ \ \ \mathbb{H}_{1st}, \mathbb{H}_{2nd}\in \mathbb{R}^{N\times H}\ or\ \mathbb{R}^{N\times t_{pred}\times H}$} & \makecell[l]{
		$\mathbf{FExM}_1$: Sparse Graph Convolution Network ($\mathcal{M}_1$) 
			\\
			$\mathbf{FExM}_2$: Multilayer Perceptron Network ($\mathcal{M}_2$)
			\\
			$\mathbf{FExM}_3$: Spatio-Temporal Graph CNN ($\mathcal{M}_3$)
			\\
			$\mathbf{FExM}_4$: LSTM based Motion Encoder Module ($\mathcal{M}_4$)
			\\
			$\mathbf{FExM}_5$: GAT-based Crowd Interaction Modeling ($\mathcal{M}_5$)
		}\\
			\cline{3-4} 	
		&&\makecell[l]{\\Feature Enhancement Method (FEnM)\\$\ \ \ \  \mathbb{H}^{\prime}_{1st} =\mathbf{FEnM}{\mathbf{1st}}(\mathbb{I}^\prime,\mathbb{H}_{1st}) 
			$\\
			$\ \ \ \ \mathbb{H}^{\prime}_{2nd} =\mathbf{FEnM}{\mathbf{2nd}}(\mathbb{I}^\prime,\mathbb{H}_{2nd}) 
			$\\
			$\ \ \ \ \mathbb{H}_{1st}^{\prime}, \mathbb{H}_{2nd}^{\prime}\in \mathbb{R}^{N\times H}\ or\ \mathbb{R}^{N\times t_{pred}\times H}$} &   \makecell[l]{ $\mathbf{FEnM}_1$: None ($\mathcal{M}_1,\mathcal{M}_3$)
			\\
			$\mathbf{FEnM}_2$: Latent Belief Energy-based Module ($\mathcal{M}_2$)
			\\
			$\mathbf{FEnM}_3$: Attention Pooling Module ($\mathcal{M}_4$)
			\\
			$\mathbf{FEnM}_4$: LSTM-based Temporal Correlation Modeling ($\mathcal{M}_5$) 
			} \\
		\cline{3-4} 	
		&& \makecell[l]{Feature Fusion Method (FFM)\\$\ \ \ \  \mathbb{H}_{\mathbf{all}}=\mathbf{FFM}(\mathbb{H}_{1st},\mathbb{H}_{2nd},{\mathbb{H}^{\prime}_{1st}}, {\mathbb{H}^{\prime}_{2nd}})$\\
		$\ \ \ \ \mathbb{H}_{\mathbf{all}} \in \mathbb{R}^{N\times H_{all}}$} &\makecell[l]{
			$\mathbf{FFM}_1$: Concentrate All Features 
			\\
			$\mathbf{FFM}_2$: Concentrate All Enhanced Features
		}\\
		\cline{3-4} 	
		&&\makecell[l]{Output Module ($\mathbf{OM}$)\\ $\ \ \ \  \mathbb{O}'=\mathbf{OM}(\mathbb{H}_{\mathbf{all}})$\\
		$\ \ \ \ \mathbb{O}' \in \mathbb{R}^{N\times t_{obs}\times 2}$}&  \makecell[l]{$\mathbf{OM}_1$: Time Convolution Network ($\mathcal{M}_1$) 
		\\
		$\mathbf{OM}_2$: Time-Extrapolator Convolution Neural Network ($\mathcal{M}_3$)
		\\
		$\mathbf{OM}_3$: Multiple Fully-Connected Layer ($\mathcal{M}_2,\mathcal{M}_4$)
		\\
		$\mathbf{OM}_4$: LSTM + Fully-Connected Layer ($\mathcal{M}_5$)} \\

		\midrule
		\textbf{Stage 2: TAD Output Content Design}
		 & Determine the output content of trajectory AD model.   &\makecell[l]{ Output Content ($\mathbf{OC}$)
			}
		&  \makecell[l]{
			$\mathbf{OC}_1$: Predicting the historical trajectory w.r.t. input $\mathbb{I}$
		}
		\\     \midrule

		\textbf{Stage 3: TAD Loss Function Design}                & Design an effective loss function for trajectory AD model training. &  \makecell[l]{Loss Function ($\mathbf{LF}$)\\$\ \ \ \ \mathcal{L}=\boldsymbol{\Lambda} \cdot \mathbf{LF}(\mathbb{O},{\mathbb{O}^{\prime}}, \mathbb{H}_{\mathbf{all}})\in \mathbb{R}$\\
		$\ \ \ \ \boldsymbol{\Lambda}=(\boldsymbol{\lambda_{1}},\boldsymbol{\lambda_{2}},\boldsymbol{\lambda_{3}},\boldsymbol{\lambda_{4}},\boldsymbol{\lambda_{5}},\boldsymbol{\lambda_{6}},\boldsymbol{\lambda_{7,1}},\boldsymbol{\lambda_{7,2}})$\\
		$\ \ \ \ \mathbf{LF}=(\mathcal{L}_{out},\mathcal{L}_{adv},\mathcal{L}_{fea},\mathcal{L}_{com},$\\$\ \ \ \ \ \ \ \ \ \ \ \ \mathcal{L}_{sep},\mathcal{L}_{clu},\mathcal{L}_{rsr}^{1},\mathcal{L}_{rsr}^{2})$} &\makecell[l]{\\$\boldsymbol{\lambda_1}$: 0.1, 0.01, 1
			\\
			$\boldsymbol{\lambda_i\ (i=2,\ldots,6)}$: 0, 0.1, 0.01, 1
			\\
			$\boldsymbol{\lambda_{7,j}\ (j=1,2)}$: 0, 0.1, 0.01, 1
		} \\     \midrule

		\multirow{2}{\linewidth}{\textbf{Stage 4: TAD Anomaly Scoring Function Design}} & \multirow{2}{\linewidth}{Design an effective anomaly scoring function to examin the rationality of the given input.}   & \makecell[l]{Scoring Function ($\mathbf{SF}$) \\$\ \ \ \ \vec{\mathcal{F}}=\boldsymbol{\Gamma} \cdot \vec{\mathbf{SF}}(\mathbb{O},{\mathbb{O}^{\prime}}, \mathbb{H}_{\mathbf{all}})\in \mathbb{R}^{N}$\\
		$\ \ \ \ \boldsymbol{\Gamma}=(\boldsymbol{\gamma_{1}},\boldsymbol{\gamma_{2}},\boldsymbol{\gamma_{3}},\boldsymbol{\gamma_{4}},\boldsymbol{\gamma_{5}},\boldsymbol{\gamma_{6}},\boldsymbol{\gamma_{7,1}},\boldsymbol{\gamma_{7,2}})$\\
		$\ \ \ \ \vec{\mathbf{SF}}=(\vec{\mathcal{L}_{out}},\vec{\mathcal{L}_{adv}},\vec{\mathcal{L}_{fea}},\vec{\mathcal{L}_{com}},$\\$\ \ \ \ \ \ \ \ \ \ \ \ \vec{\mathcal{L}_{sep}},\vec{\mathcal{L}_{clu}},\vec{\mathcal{L}_{rsr}^{1}},\vec{\mathcal{L}_{rsr}^{2}})$
		}
		&\makecell[l]{\\$\boldsymbol{\gamma_i\ (i=1,\ldots,6)}$: 0, $\boldsymbol{\lambda_{i}}$\\ $\boldsymbol{\gamma_{7,j}\ (j=1,2)}$: 0, $\boldsymbol{\lambda_{7,j}}$
		}
		\\     
	\bottomrule
	\end{tabular}
	}
	\label{table3}
\end{table*}

\begin{figure*}
\begin{center}
\includegraphics[width=1.0\textwidth]{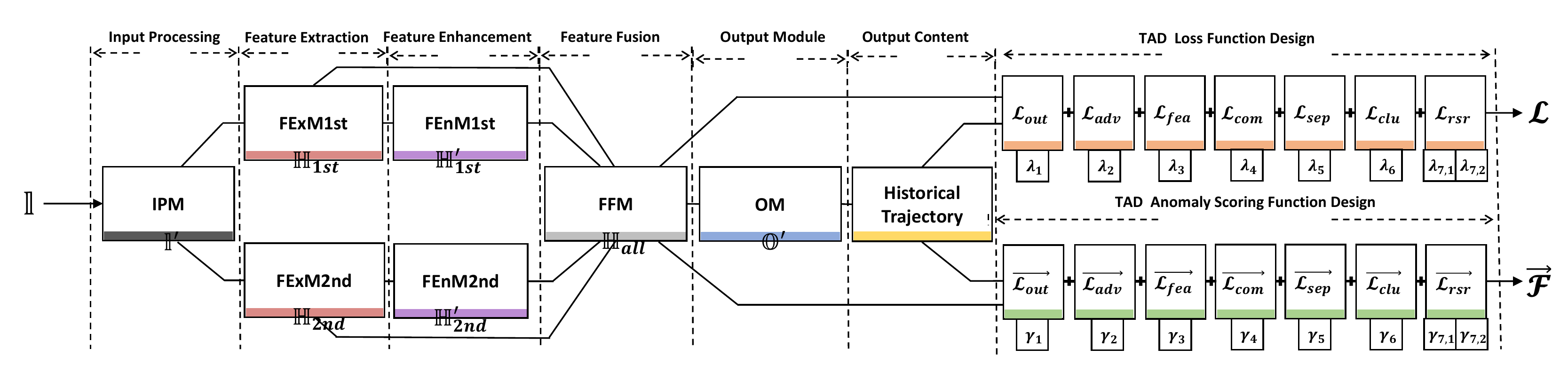}
\end{center}
\caption{Details of each trajectory AD model in the search space $\mathbb{S}$.}
\label{fig1}
\end{figure*}

\textbf{*Stage 1: TAD Model Architecture Design.} According to analysis on Stage 1 of the existing AD models, the architecture of the trajectory AD model should be suitable for the human trajectory data. The existing TP works provide many effective modules to analyze and predict trajectory data, and we can utilize these modules to construct the trajectory AD model.

Specifically, the existing TP models are composed of five parts: Input Processing Module ($\mathbf{IPM}$),  Feature Extraction Module ($\mathbf{FExM}$), Feature Enhancement Module ($\mathbf{FEnM}$), Feature Fusion Module ($\mathbf{FFM}$) and Output Module ($\mathbf{OM}$). These modules cover all operations required by the AD model architecture. In TPAD, we can extract these moudles from existing TP models and thus construct an effective search space for trajectory AD model architecture design.

\textbf{*Stage 2: TAD Output Content Design.} In TPAD, incorrect trajectory predictions generated by the TP model are considered as abnormalities. These abnormalities may be slightly different from the normal ones, but are difficult to accurately map to the corresponding historical trajectories, due to violation of mapping rules. 

According to analysis on Stage 2 of the existing AD models, this kind of AD task is fit for predictive AD models. Therefore, in TPAD, we take the known historical trajectory w.r.t. the input as the target output of the trajectory AD model. We aim to use the trajectory AD model to learn mapping rules from real future trajectories and historical trajectories, so as to reasonably identify trajectory prediction results with low precision.

\textbf{*Stage 3: TAD Loss Function Design.} According to analysis on Stage 3 of the existing AD models, the combination of multiple loss functions may improve the training effect of the AD model. Therefore, in TPAD, we allow different loss functions designed for the existing AD models (in Table~\ref{table2}) to be combined, so as to construct more effective loss functions to guide the trajectory AD model training.

\textbf{*Stage 4: TAD Anomaly Scoring Function Design.} According to analysis on Stage 4 of the existing AD models, the anomaly scoring function is generally designed based on the loss function of the AD model. Therefore, in TPAD, we will construct the anomaly scoring function using components of the loss function designed for the trajectory AD model.

\subsection{Search Space Design}\label{section:3.2}
Following the design scheme obtained in Section~\ref{section:3.1}, we design an effective search space $\mathbb{S}$ on trajectory AD model. Table~\ref{table3} gives the components of the search space $\mathbb{S}$ and Figure~\ref{fig1} shows the details of each trajectory AD model in the search space $\mathbb{S}$.

In this search space, we collect diversified architecture components from existing TP models, allowing components of different sources to be combined to construct more flexible trajectory AD model architecture. Note that we use two groups of feature analysis components, i.e., $\mathbf{(FExM1st, FEnM1st)}$ and $\mathbf{(FExM2nd, FEnM2nd)}$, for architecture construction (as is shown in Figure~\ref{fig1}), so as to improve the fitting ability of the trajectory AD models.

In addition, we extract effective loss functions from the different AD works to guide the trajectory AD model training and scoring function construction. We allow these functions to have different weights and jointly guide model to gain more valuable trajectory AD criterions. Note that different from the traditional real-valued anomaly scoring functions, we preserve the anomaly score of each pedestrian constructing a $N$-dimensional\footnote{$N$ is the number of pedestrians in the given trajectory scene.} anomaly score in TPAD. This anomaly scoring function can guide TPAD to select effective trajectory prediction result for each pedestrian, achieving better prediction effect.

In all, our search space $\mathbb{S}$ contains $9.6\times 10^3$ trajectory AD model architectures and about $4.9\times 10^4$ anomaly loss and scoring functions. These materials constructs diversified trajectory AD model design schemes, can support the realization of our automatic trajectory AD model design.

\subsection{Search Strategy Selection}\label{section:3.3}

The search space $\mathbb{S}$ designed in Section~\ref{section:3.2} is huge. In this part, we aim to analyze the existing AutoML techniques and find effective search strategy to guide the efficient search of the suitable trajectory AD model.

\textbf{Existing AutoML Techniques.} The search methods used in existing AutoML algorithms can be classified in three categories, i.e., RL-based methods~\cite{DBLP:conf/ijcai/GaoYZ0H20, DBLP:conf/icml/BelloZVL17}, EA-based methods~\cite{DBLP:conf/aaai/RealAHL19,DBLP:conf/cvpr/ChenMZXHMW19} and gradient based methods~\cite{DBLP:journals/pr/Tian0XJY21,DBLP:conf/cvpr/HosseiniYX21}.

EA-based methods are generally used for solving multi-objective AutoML problems. They can effectively search pareto optimal solutions, but are less effective than the other two search methods in single-objective AutoML problems. In TPAD, we aim to find a trajectory AD model with the highest $\mathbf{AUC}$ score, and therefore do not apply the EA-based search methods. As for the gradient based methods, they require the loss function to be fixed and alternately optimize model weights and model details by minimizing it. However, our trajectory AD model search space contains diversified loss functions (as is shown in Stage 3 of Table~\ref{table3}). 
The applied loss function will change with the trajectory AD model design scheme, and this can make optimization process unstable and ineffective.

RL-based search methods are widely used for single-objective problems and have no special requirements for the search space. Therefore, in TPAD, we choose to use RL method to search the optimal trajectory AD model from the search space $\mathbb{S}$.

\begin{figure}
\begin{center}
\includegraphics[width=1.0\columnwidth]{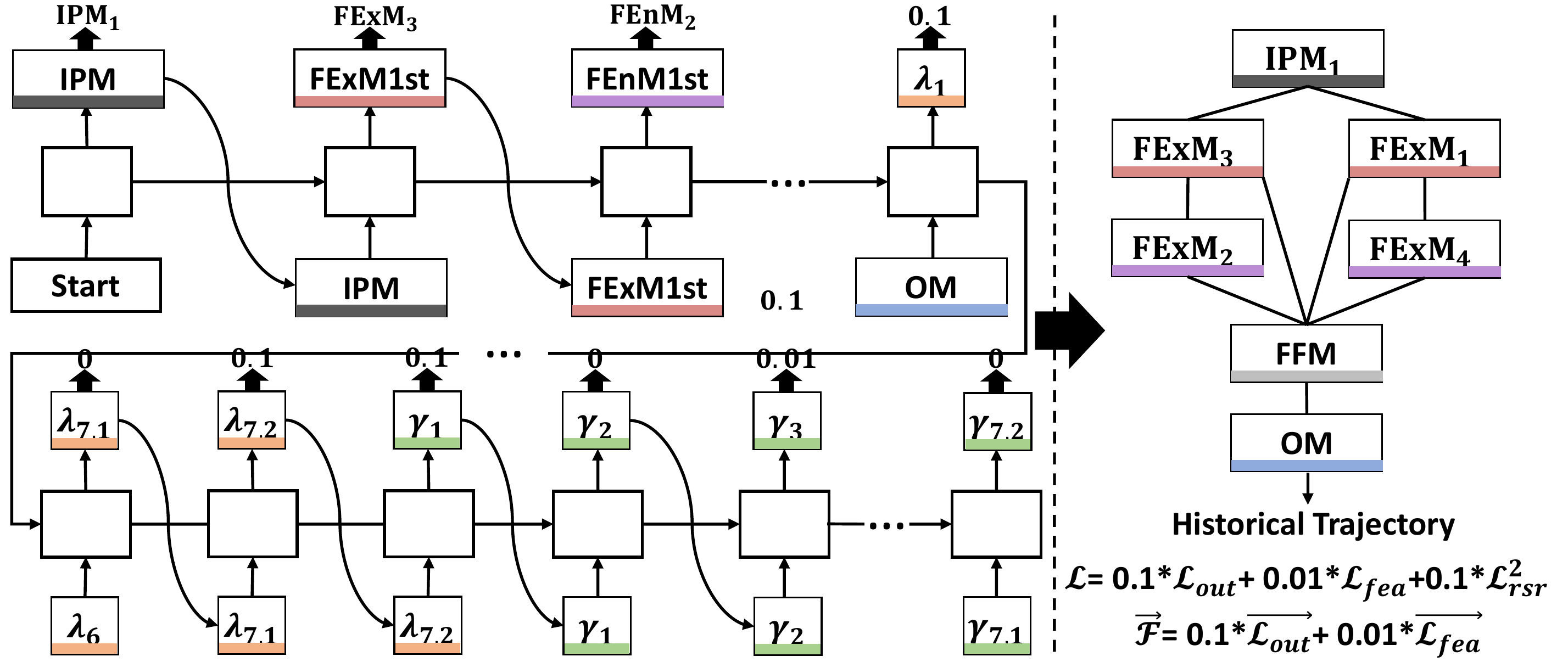}
\end{center}
\caption{An illustration of the RNN controller constructing a trajectory AD model.}
\label{fig4}
\end{figure}

\textbf{RL-based Trajectory AD model Search.} In TPAD, we use a RNN model parameterized by $\theta$ as controller to determine a sequence of operators with length 23, denoted by $\mathcal{O}_{1:23}$. Each operator $\mathcal{O}_{i}\ (i=1,\ldots,23)$ corresponds to a block in Figure~\ref{fig1}, and is sampled from the corresponding options in Table~\ref{table3}. Therefore, the RNN controller can construct a trajectory AD model in the search space $\mathbb{S}$ sequentially. Figure~\ref{fig4} gives an example of constructing a trajectory AD model by the RNN controller. After the controller generates an operator sequence $\mathcal{O}_{1:23}$, we build a trajectory AD model $\mathcal{F}_{\mathcal{O}_{1:23}}\in \mathbb{S}$. We use the normal training trajectories $\mathbb{D}_{train}$ to train $\mathcal{F}_{\mathcal{O}_{1:23}}$, and use validation trajectories combined with negative samples $\mathbb{D}_{val}\cup \mathbb{D}_{val}^{-}$ to examine the $\mathbf{AUC}$ score of $\mathcal{F}_{\mathcal{O}_{1:23}}$. 

To make the controller generates a better trajectory AD model (with higher $\mathbf{AUC}$ score) over time, we use $\mathbf{AUC}$ score as a reward signal $\mathcal{R}$ and apply the policy gradient algorithm to update parameters $\theta$.  Since the reward signal $\mathcal{R}$ is non-differentiable, in TPAD, we iteratively update $\theta$ using REINFORCE~\cite{Williams1992Simple} as follows.
\begin{equation}
\begin{split}
&\nabla_{\theta} \mathbb{E}_{P(\mathcal{O}_{1:23};\theta)}[\mathcal{R}]\\
&=\sum_{t=1}^{23} \mathbb{E}_{P(\mathcal{O}_{1:23};\theta)}\big[\nabla_{\theta}logP(\mathcal{O}_{t}|\mathcal{O}_{t-1:1};\theta)(\mathcal{R}-b)\big]
\end{split}
\label{equ1}
\end{equation}
where $b$ is an exponential moving average of the previous trajectory AD model rewards. With the increase of historical evlaution information, the RNN controller becomes more effective, and thus be able to guide TPAD to find a high-performance trajectory AD model.

\begin{figure*}
\begin{center}
\includegraphics[width=1.0\textwidth]{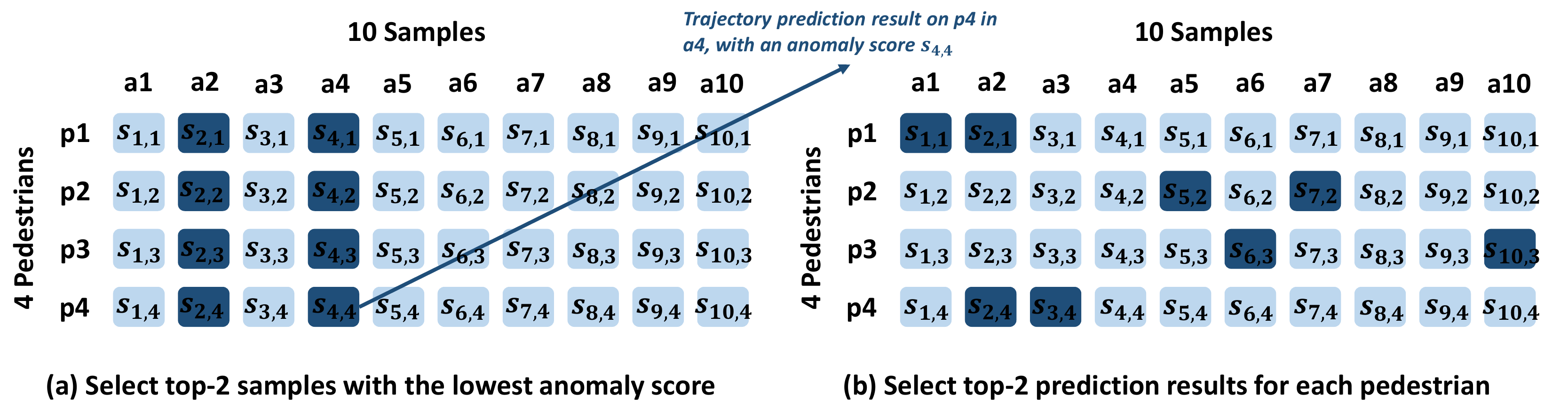}
\end{center}
\caption{Suppose there are 4 pedestrians in the scene and a stochastic TP model generate 10 samples for their future trajectories. Each cell denotes the trajectory prediction result on a certain pedestrian $pi$ in a sample $aj$, and corresponds a anomaly score $s_{i,j}$. TPAD can flexible assemble better samples by selecting top prediction results (with the lowest anomaly score) from different samples for each pedestrian.}
\label{fig2}
\end{figure*}

\subsection{TPAD Method}\label{section:3.4}

In TPAD, we use the obtained optimal trajectory AD model (denoted  by $\mathcal{F}^{*}$) to help stochastic TP models to identify reasonable trajectory predictions. Suppose a stochastic TP model $\mathcal{M}$ generates $\Psi$ samples, i.e., $\Psi$ future trajectory prediction results: $\widehat{\mathbb{Y}_{\mathcal{M},i}}\in \mathbb{R}^{N \times t_{pred}\times 2}\ (i=1,\ldots,\Psi)$ on the given input trajectory $\mathbb{X}\in \mathbb{R}^{N \times t_{obs}\times 2}$, we aim to utilize $\mathcal{F}^{*}$ to filter out $\psi \ll \Psi$ high-quality prediction results for each pedestrian. In this way, the stochastic TP model can provide fewer but better prediction results in real applications, achieving higher availability and reliability. 

As mentioned in Section~\ref{section:3.0}, given a future trajectory prediction result $\widehat{\mathbb{Y}_{\mathcal{M},i}}\in \mathbb{R}^{N \times t_{pred}\times 2}$, the trajectory AD model $\mathcal{F}^{*}$ will return a $N$-dimensional anomaly score: $\mathcal{F}^{*}(\widehat{\mathbb{Y}_{\mathcal{M},i}}, \mathbb{X})\in \mathbb{R}^{N}$. Each dimension corresponds to an anomaly value on a certain pedestrian, examining the rationality of the predicted future trajectories w.r.t. this pedestrian. $\Psi$ samples will generate a $N \times \Psi$ anomaly score matrix, where $N$ denotes the number of pedestrians in the given trajectory scene. Figure~\ref{fig2} gives an example. With known the anomaly score of each pedestrian's prediction result in each sample, TPAD can select top $\psi$ prediction results (with the lowest anomaly score) for each pedestrian separately from the anomaly score matrix. This can help $\mathcal{F}^{*}$ to assemble more effective prediction results, playing out the high performance.

Note that different from the traditional AD works, we do not make the trajectory AD model output an anomaly value for the given input, but to examine the abnormal degree of each pedestrian's trajectory seperately. Our multidimensional anomaly scoring method is more flexible, and more suitable for TP area. It can fusion the benefits of different samples, and thus improve the TPAD's performance. Algorithm~\ref{alg1} gives the pseudo code of TPAD.

\begin{algorithm}[t]
	\caption{TPAD Agorithm}
		\label{alg1}
		\textbf{Input:} A TP dataset $\mathbb{D}=\{\mathbb{D}_{train}, \mathbb{D}_{val}\}$, a stochastic TP model $\mathcal{M}$, a historical trajectory scene$\mathbb{X}\in \mathbb{R}^{N\times t_{obs}\times 2}$, sample number $\Psi$\\
		\textbf{Output:} Top $\psi$ prediction results of each pedestrian in the given scene $\widehat{\mathbb{Y}_{\mathcal{M},i}^{p*}}$ ($i$=1,...,$\psi$;p=1,...,N)
		\begin{algorithmic}[1]
		\STATE \# Initialization
		\STATE $\mathbb{S}\gets$ Construct a search space on trajectory AD model according to Table~\ref{table3} and Figure~\ref{fig1}
		\STATE $RNN_{\theta}\gets$ Construct a RNN controller according to Figure~\ref{fig4}
		\STATE $\mathbb{D}_{val}^{-}\gets$ Add noises to validation set $\mathbb{D}_{val}$ constructing negative samples
		\STATE \# Design an Optimal Trajectory AD Model for $\mathbb{D}$
		\WHILE {epoch $<$ $SearchEpoch$}
			\STATE $\mathcal{O}_{1:23}\gets$ generate an opoerator sequence using $RNN_{\theta}$
			\STATE $\mathcal{F}\gets$ using operators $\mathcal{O}_{1:23}$ to construct a trajectory AD model in $\mathbb{S}$
			\STATE $\mathcal{R}\gets$ train $\mathcal{F}$ using $\mathbb{D}_{train}$ and calculate its $\mathbf{AUC}$ score using $\mathbb{D}_{val}\cup \mathbb{D}_{val}^{-}$
			\STATE Update controller's parameter $\theta$ according to \ref{equ1}
		\ENDWHILE
		\STATE $\mathcal{F}^{*}\gets$ Optimal trajectory AD model (with the highest reward score $\mathcal{R}$) generated by $RNN_{\theta}$
		\STATE \# Identify Effective Prediction Results
		\STATE $\widehat{\mathbb{Y}_{s}}\gets$ $\{\widehat{\mathbb{Y}_{\mathcal{M},i}}=\mathcal{M}(\mathbb{X}) | i = 1,\ldots,\Psi\}$
		\STATE $AnomScores\gets \{\mathcal{F}^{*}(\widehat{\mathbb{Y}_{\mathcal{M},i}},\mathbb{X})\in \mathbb{R}^{N} | \widehat{\mathbb{Y}_{\mathcal{M},i}}\in \widehat{\mathbb{Y}_{s}}\}$
		\STATE $\widehat{\mathbb{Y}_{\mathcal{M},i}^{p*}}, \ldots, \widehat{\mathbb{Y}_{\mathcal{M},\psi}^{p*}}\gets$ Select $\psi$ results with lowest anomaly score for pedestrian $p$
		\RETURN $\widehat{\mathbb{Y}_{\mathcal{M},i}^{p*}}, \ldots, \widehat{\mathbb{Y}_{\mathcal{M},\psi}^{p*}}\ (p=1,\ldots,N)$ 
	    \end{algorithmic}
\end{algorithm}

\section{Experiments}\label{section:4}
In this section, we analyze the performance of TPAD and demonstrate its significance in the TP area. Our experiments are divided into three parts:
\begin{itemize}
\item[1.] Firstly, we compare the optimal trajectory AD model $\mathcal{F}^{*}$ searched by TPAD with the manually designed ones, analyzing the importance of AutoML in TPAD and the rationality of our search space design (Section~\ref{section:4.1}). \\
\item[2.] Secondly, we analyze the quality of multiple prediction results provided by state-of-the-arts stochastic TP models, and examine the ability of TPAD in identifying high-quality trajectory prediction results (Section~\ref{section:4.2}).
\item[3.] Finally, we analyze the effect of the sample number of stochastic TP models and the number of filtered prediction results on the performance of TPAD (Section~\ref{section:4.3}).
\end{itemize}
We implemented all algorithms using Pytorch and performed all experiments using RTX 3090 GPUs.

\begin{figure*}[t]
\includegraphics[scale=0.5]{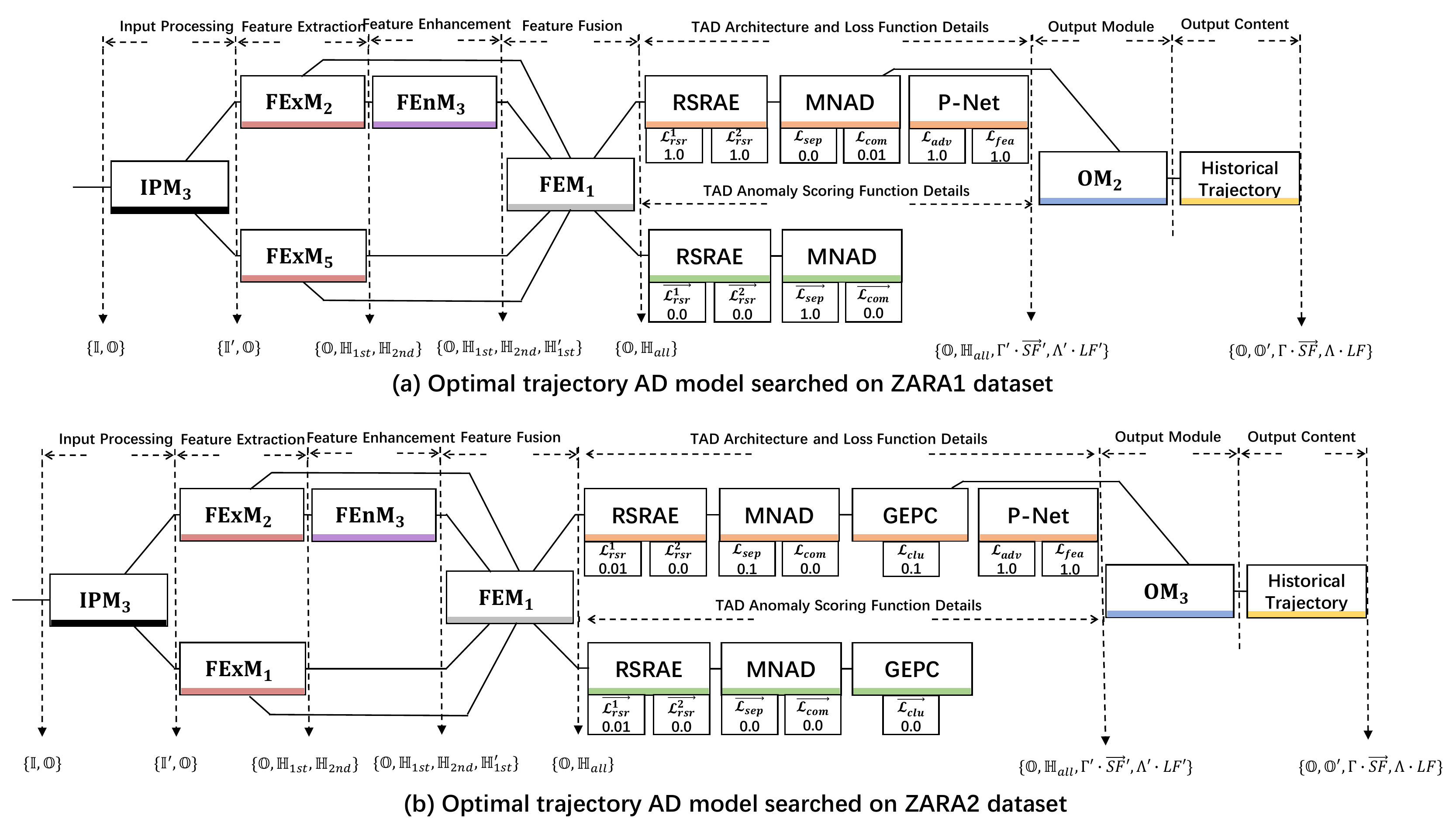}
\caption{The optimal trajectory AD model searched by the RL-based AutoML method in TPAD. As can be seen from the figure, the TP related modules and AD strategies coming from different sources can be flexible combined in TPAD. Note that the weight of each loss value (or vector) applied in the trajectory AD model is shown below the corresponding symbol.}
\label{fig:searched-model}
\end{figure*}

\subsection{Experimental Setup}\label{section:4.0}
\paragraph{\textbf{Datasets.}} We use two public pedestrian trajectory datasets, i.e., ETH~\cite{DBLP:journals/pr/0007PPF21} and UCY~\cite{lerner2007crowds}, to analyze stochastic TP models and examine the performance of our proposed method. Two datasets consist of five different trajectory scenes, i.e., ETH dataset contains the ETH and HOTEL scenes, while the UCY dataset contains UNIV, ZARA1, and ZARA2 scenes, all of which provide pedestrian positions in world coordinates. 

In the experiments, we follow the ``leave one out'' method \cite{DBLP:journals/pr/ZamboniKGNC22} in previous works for TP model and trajectory AD model training and evaluation. Each stochastic TP model observes 8 frames (3.2 seconds) trajectories and predicts the next 12 frames (4.8 seconds). As for the trajectory AD models involved in TPAD, they take the 12-frame future trajectories as input, and predict 8 historical frames observed by TP models.

Note that trajectory AD models need both positive and negative trajectory samples for performance evaluation. We generate negative samples by randomly adding noises ranging from -0.1 to 0.1 to the ground-truth future trajectories. We use the normal pedestrian trajectories provided by the dataset combined with our manually generated negative samples to calculate the $\mathbf{AUC}$ scores of the trajectory AD models. 

\paragraph{\textbf{Baselines or Compared Algorithms.}} To examine significance of the AutoML part in the TPAD, we compare the trajectory AD model searched by TPAD with some manually designed ones. We use SGCN's~\cite{DBLP:conf/cvpr/Shi0LZZN021} TP model architecture, and follow the loss function and anomaly scoring function designed by 4 stat-of-the-arts AD methods: MNAD~\cite{park2020learning}, P-NET~\cite{DBLP:conf/eccv/ZhouXYCLLGLG20}, RSRAE~\cite{DBLP:conf/iclr/LaiZL20}, and GEPC~\cite{markovitz2020graph}, and thus construct some trajectory AD models. We also take the random search as a baseline, to examine the effectiveness of RL-based search strategy applied in the TPAD.

In addition, we analyze 4 state-of-the-arts stochastic TP models, including Social-STGCNN~\cite{mohamed2020social}, STAR~\cite{DBLP:conf/eccv/YuMRZY20}, STGAT~\cite{huang2019stgat} and SGCN~\cite{DBLP:conf/cvpr/Shi0LZZN021}. Four models apply different methods to generate diversified trajectories, i.e., Social-STGCNN and SGCN model the pedestrian trajectories as a bi-variate Gaussian distribution, while STAR and STGAT add the random Gaussian noise to their predictions. We compare the performance of these 4 stochastic TP models before and after applying TPAD, so as to examine the effectiveness of our proposed algorithm.

\setlength{\tabcolsep}{9mm}{
\begin{table*}
\caption{
The RL-based search method in TPAD vs. random search and the manually designed trajectory AD models. We report the $\mathbf{AUC}$ score of each trajectory AD model.
}
\centering
\resizebox{\linewidth}{!}{ 
\begin{tabular}{@{}l|ccccc|c@{}}
\toprule
Method & ETH & HOTEL & ZARA1 & ZARA2 & UNIV & Average\\
\midrule
SGCN+RSRAE & 0.540 & 0.524 & 0.537 & 0.522 & 0.527 & 0.530 \\
SGCN+GEPC & 0.507 & 0.516 & 0.512 & 0.512 & 0.516 & 0.513 \\
SGCN+MNAD & 0.508 & 0.514 & 0.510 & 0.511 & 0.511 & 0.516 \\
SGCN+MNAD+GEPC & 0.508 & 0.513 & 0.510 & 0.513 & 0.512 & 0.511 \\
SGCN+P-Net & 0.507 & 0.516 & 0.509 & 0.510 & 0.511 & 0.511 \\
\midrule
Random Search & 0.635 & \textbf{0.760} & 0.777 & \textbf{0.802} & 0.712 & 0.737 \\
\textbf{TPAD} & \textbf{0.792} & 0.752 &  \textbf{0.816} & 0.731 & \textbf{0.840} & \textbf{0.786}\\
\bottomrule
\end{tabular}
} 
\label{tab:compare-automl}
\end{table*}
}

\begin{figure*}
\begin{center}
\includegraphics[width=2.0\columnwidth]{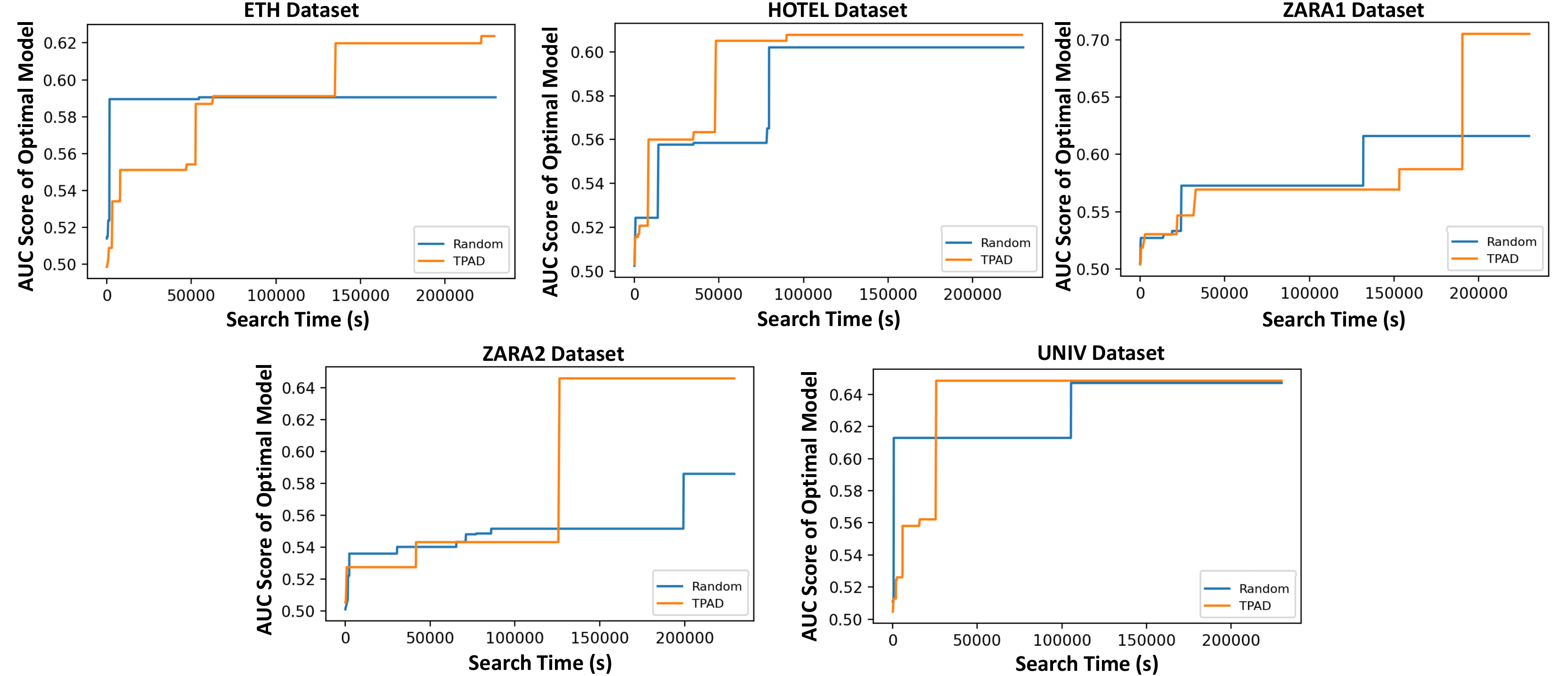}
\end{center}
\caption{RL-based search method in TPAD vs. Random Search method. Our proposed RL-based search method can obtain a better trajectory AD model with higher $\textbf{AUC}$ value faster than the random method.
}
\label{fig:search_comparison}
\end{figure*}

\paragraph{\textbf{Evaluation Metrics.}} For the trajectory AD models searched in the AutoML part of TPAD, we use them to calculate anomaly scores of the negative and positive validation trajectory samples, and apply Area Under ROC Curve ($\mathbf{AUC}$) to examine their effectiveness. 

As for the stochastic TP models, we make them to generate $\Psi$ trajectory prediction samples ($\Psi=50$ by default) for each input. Then, we employ two metrics, namely Average Displacement Error ($\mathbf{ADE}$)~\cite{alahi2016social} and Final Displacement Error ($\mathbf{FDE}$)~\cite{DBLP:journals/pr/BarataNLM21,DBLP:conf/cvpr/PangZ0W21} to evaluate their prediction results. We report the average score of these samples (denoted by $\mathbf{Average}$), the score of the best prediction result assembled by these samples (denoted by $\mathbf{Best}$), and the score of the worst sample (denoted by $\mathbf{Worst}$), so as to make comprehensive analysis of the practical effect of the stochastic TP models. 

In the experiments, we use TPAD algorithm to guide stochastic TP models to select Top-$\psi$ ($\psi=10$ by default) predictions results for each pedestrian in the given input. We report the same performance scores, i.e., $\mathbf{Average}$, $\mathbf{Best}$ and $\mathbf{Worst}$, using the filtered samples, so as to analyze the effectiveness of TPAD.

\paragraph{\textbf{Implementation Details.}} In TPAD, the embedding size and hidden size are set to 100. RNN controller is trained with the Adam optimizer with a learning rate of
3.5e-4. For each trajectory AD model candidate, we train it for 3
epochs. After the controller searches for 3 GPU days, we choose the best trajectory AD model $\mathcal{F}^{*}$ that achieves the highest $\mathbf{AUC}$ score, and train it for 50 epochs. Figure~\ref{fig:searched-model} shows details of the trajectory AD model searched on ZARA1 and ZARA2.

\setlength{\tabcolsep}{2.7mm}{
\begin{table*}
\caption{
Performance comparison of deterministic TP models, stochastic TP models and tochastic TP models that applied TPAD method.
} 
\newcommand{\tabincell}[2]{\begin{tabular}{@{}#1@{}}#2\end{tabular}}
\centering
\resizebox{\linewidth}{!}{ 
\begin{tabular}{@{}l|ll|ccccc|l@{}}
\toprule
\multirow{2}{*}{\textbf{Type}} & \multicolumn{2}{l|}{\multirow{2}{*}{\textbf{TP Models}}} & \multicolumn{5}{c|}{\textbf{Datasets}} & \multirow{2}{*}{\textbf{Average}} \\
\cline{4-8}
&  & & \textbf{ETH} & \textbf{HOTEL} & \textbf{ZARA1} & \textbf{ZARA2} & \textbf{UNIV} & \\
\toprule
\multirow{5}{*}{\tabincell{l}{Deterministic \\TP Models}} & \multicolumn{2}{l|}{S-LSTM~\cite{alahi2016social}} & \textbf{0.77 / 1.60} & 0.38 / 0.80 & 0.51 / 1.19 & 0.39 / 0.89  & \textbf{0.58 / 1.28} & 0.53 / 1.15 \\
& \multicolumn{2}{l|}{CIDNN~\cite{DBLP:conf/cvpr/XuPG18}} & 1.25 / 2.32 & 1.31 / 1.86 & 0.90 / 1.28 & 0.50 / 1.04 & 0.51 / 1.07 & 0.89 / 1.73 \\
& \multicolumn{2}{l|}{SocialAttention~\cite{DBLP:conf/icra/VemulaMO18}} & 1.39 / 2.39  & 2.51 / 2.91 & 1.25 / 2.54 & 1.01 / 2.17 & 0.88 / 1.75 & 1.41 / 2.35 \\
& \multicolumn{2}{l|}{TrafficPredict~\cite{2018TrafficPredict}} & 5.46 / 9.73 & 2.55 / 3.57  & 4.32 / 8.00 & 3.76 / 7.20 & 3.31 / 6.37 & 3.88 / 6.97 \\
& \multicolumn{2}{l|}{PITF~\cite{DBLP:conf/cvpr/Liang0NH019}} & 0.88 / 1.98 & \textbf{0.36 / 0.74} & \textbf{0.42 / 0.90} & \textbf{0.34 / 0.75} & 0.62 / 1.32 & \textbf{0.52 / 1.14} \\
\bottomrule

\multirow{20}{*}{\tabincell{l}{Stochastic\\ TP Models}} & \multirow{5}{*}{Social-STGCNN} & $\mathbf{Best}$ & 0.59 / 0.93 & 0.31 / 0.46 & 0.31 / 0.45 & 0.27 / 0.40 & 0.40 / 0.64 & 0.37 / 0.57 \\
& & $\mathbf{Best}$ (TPAD Top-10) & 0.77 / 1.35 & 0.37 / 0.62 & 0.39 / 0.66 & 0.35 / 0.56 & 0.50 / 0.88 & 0.47 / 0.81 \\
\cline{3-9}
& & $\mathbf{Average}$ & 1.23 / 2.22 & 0.73 / 1.37 & 0.68 / 1.25 & 0.59 / 1.08 & 0.83 / 1.52 & 0.81 / 1.48 \\
& & $\mathbf{Average}$ (TPAD Top-10) & 1.20 / 2.20 & 0.65 / 1.22 & 0.66 / 1.23 & 0.58 / 1.07 & 0.82 / 1.51 & 0.78 / 1.44 \\
\cline{3-9}
& & $\mathbf{Worst}$ & 1.88 / 3.02 & 1.04 / 1.87 & 0.94 / 1.65 & 0.76 / 1.34 & 0.93 / 1.69 & 1.11 / 1.91 \\
\cline{2-9}

 & \multirow{5}{*}{STGAT} & $\mathbf{Best}$ & 0.52 / 0.88 & \textbf{0.19 / 0.30} & \textbf{0.16 / 0.27} & \textbf{0.15 / 0.26} & \textbf{0.22 / 0.39} & \textbf{0.24 / 0.42} \\
& & $\mathbf{Best}$ (TPAD Top-10) & 0.71 / 1.37 & \textbf{0.23 / 0.40} & \textbf{0.26 / 0.52} & 0.28 / 0.58 & \textbf{0.35 / 0.71} & \textbf{0.36 / 0.69} \\
\cline{3-9}
& & $\mathbf{Average}$ & 1.24 / 2.63 & 0.70 / 1.49 & 0.73 / 1.62 & 0.58 / 1.32 & 0.82 / 1.80 & 0.81 / 1.77 \\
& & $\mathbf{Average}$ (TPAD Top-10) & 1.09 / 2.28 & \textbf{0.43 / 0.85} & 0.51 / 1.09 & 0.42 / 0.92 & 0.66 / 1.44 & 0.62 / 1.31 \\
\cline{3-9}
& & $\mathbf{Worst}$ & 1.39 / 2.95 & 0.82 / 1.76 & 0.84 / 1.87 & 0.69 / 1.58 & 0.88 / 1.93 & 0.92 / 2.01 \\
\cline{2-9}

 & \multirow{5}{*}{SGCN} & $\mathbf{Best}$ & 0.57 / 0.78& 0.24 / 0.39 & 0.25 / 0.42 & 0.21 / 0.36 & 0.32 / 0.54 & 0.31 / 0.49  \\
& & $\mathbf{Best}$ (TPAD Top-10) & 0.72 / 1.22 & 0.28 / 0.48 & 0.32 / 0.58 & \textbf{0.25 / 0.49} & 0.40 / 0.74 & 0.39 / 0.70 \\
\cline{3-9}
& & $\mathbf{Average}$ & 1.12 / 2.34 & \textbf{0.49 / 1.00} & \textbf{0.48 / 1.00} & \textbf{0.38 / 0.80} & \textbf{0.61 / 1.27} & \textbf{0.61 / 1.28} \\
& & $\mathbf{Average}$ (TPAD Top-10) & 1.11 / 2.31 & 0.44 / 0.86 & \textbf{0.48 / 0.99} & \textbf{0.38 / 0.80} & \textbf{0.61 / 1.26} & \textbf{0.60 / 1.24} \\
\cline{3-9}
& & $\mathbf{Worst}$ & 1.61 / 3.47 & \textbf{0.68 / 1.39} & \textbf{0.63 / 1.27} & \textbf{0.47 / 0.98} & \textbf{0.69 / 1.41} & \textbf{0.81 / 1.70} \\
\cline{2-9}

 & \multirow{5}{*}{STAR} & $\mathbf{Best}$ & \textbf{0.30 / 0.53} & 0.18 / 0.34 & 0.21 / 0.44 & 0.21 / 0.47 & 0.54 / 1.08 & 0.28 / 0.57  \\
& & $\mathbf{Best}$ (TPAD Top-10) & \textbf{0.44 / 0.83} & 0.30 / 0.65 & 0.49 / 1.11 & 0.35 / 0.77 & 0.61 / 1.24 & 0.43 / 0.92 \\
\cline{3-9}
& & $\mathbf{Average}$ & \textbf{0.70 / 1.88} & 0.43 / 1.13 & 1.03 / 2.44 & 0.69 / 1.74 & 0.97 / 1.92 & 0.76 / 1.82 \\
& & $\mathbf{Average}$ (TPAD Top-10) & \textbf{0.62 / 1.31} & 0.41 / 0.95 & 0.80 / 1.88 & 0.55 / 1.32 & 0.89 / 2.33 &  0.65 / 1.55 \\
\cline{3-9}
& & $\mathbf{Worst}$ & \textbf{1.06 / 2.27} & 0.68 / 1.84 & 2.80 / 6.31 & 1.53 / 3.94 & 3.06 / 6.99 & 1.82 / 4.27 \\
\bottomrule

\end{tabular}
} 
\label{table5}
\end{table*}
}

\subsection{Auto-Construction of Trajectory AD model}
\label{section:4.1}

We compare the optimal trajectory AD model searched by TPAD with those obtained by the manual combination of existing methods or random search, and report the $\mathbf{AUC}$ results in Table~\ref{tab:compare-automl}. From Table~\ref{tab:compare-automl}, we can observe that the performance of the manually designed trajectory AD models is poor, and the optimal trajectory AD model searched by the RL-based method in TPAD is significantly better than that searched by the random method. We analyze the reasons for these results as follow.

\textbf{AutoML Method vs. Artificial Construction.} The manually designed trajectory AD models in Table~\ref{tab:compare-automl} are the combination of the existing TP model and AD algorithm. These state-of-the-arts TP models and AD algorithms perform well in their respective field, but fail to assemble a high-performance trajectory AD model. This is because the TP models have different design target with the trajectory AD models and the existing AD algorithms are designed for different research objects. We need to select suitable design components from existing ones, and find the most effective trajectory AD model design scheme, so as to effectively deal with our newly presented trajectory AD task.

Figure~\ref{fig:searched-model} shows the details of the optimal trajectory AD models searched by TPAD. We can observe that these models use design components from different sources to construct the model architecture, and combine multiple AD strategies to execute the model training and anomaly scoring. The flexible combination of existing design components constructs more powerful trajectory AD models compared to the manually designed ones. This result also demonstrates the rationality and effectiveness of our designed trajectory AD model search space. 

\textbf{RL-based Search vs. Random Search.} We analyze the performance of the RL-based method with the random method on the trajectory AD model search. Table~\ref{tab:compare-automl} gives the $\textbf{AUC}$ score of the optimal trajectory AD model searched by them and Figure~\ref{fig:search_comparison} shows their obtained optimal AUC score versus the search time. We can observe that the RL-based search method applied in TPAD is superior to the random method on both efficiency and effectiveness.

Though two methods explore the same search space that we designed as Table~\ref{table3} shown, their performance is quite different. RL-based AutoML method can effectively analyze the historical evaluation information and thus further improve the search efficiency and quality. This result shows us the importance of applying the effective AutoML method in TPAD. The RL-based AutoML method is necessary for TPAD, it can help TPAD to find more suitable trajectory AD model and thus achieve better identification effects.

\setlength{\tabcolsep}{2.7mm}{
\begin{table*}
\caption{
Performance of stochastic TP models which apply TPAD methods with different sample numbers $\Psi$ on HOTEL dataset.
} 
\newcommand{\tabincell}[2]{\begin{tabular}{@{}#1@{}}#2\end{tabular}}
\centering
\resizebox{0.93\linewidth}{!}{ 
\begin{tabular}{@{}ll|ccccc@{}}
\toprule
\multicolumn{2}{l|}{\multirow{2}{*}{\textbf{Stochastic TP Models}}} & \multicolumn{5}{c}{\textbf{Sample Number $\Psi$}} \\
\cline{3-7}
 & & \textbf{$\Psi=20$} & \textbf{$\Psi=40$} & \textbf{$\Psi=60$} & \textbf{$\Psi=80$} & \textbf{$\Psi=100$} \\
\toprule

\multirow{5}{*}{STGAT} & $\mathbf{Best}$ & 0.21 / 0.37 & 0.19 / 0.28 & 0.17 / 0.26 & 0.16 / 0.27 & 0.16 / 0.25 \\
& $\mathbf{Best}$ (TPAD Top-10) & 0.23 / 0.41 & 0.23 / 0.41 & 0.23 / 0.40 & 0.22 / 0.41 & 0.21 / 0.36 \\
\cline{3-7}
& $\mathbf{Average}$ & 0.68 / 1.48 & 0.73 / 1.62 & 0.73 / 1.56& 0.75 / 1.63 & 0.75 / 1.63 \\
& $\mathbf{Average}$ (TPAD Top-10) & 0.46 / 0.96 & 0.42 / 0.82 & 0.42 / 0.81 & 0.39 / 0.77 & 0.40 / 0.78 \\
\cline{3-7}
& $\mathbf{Worst}$ & 0.79 / 1.73 & 0.84 / 1.87 & 0.83 / 1.80 & 0.86 / 1.90 & 0.89 / 1.96 \\
\cline{2-7}

\multirow{5}{*}{SGCN} & $\mathbf{Best}$ & 0.26 / 0.37 & 0.24 / 0.34 & 0.23 / 0.30 & 0.22 / 0.27 & 0.21 / 0.29 \\
& $\mathbf{Best}$ (TPAD Top-10) & 0.28 / 0.48 & 0.28 / 0.45 & 0.28 / 0.43 & 0.26 / 0.36 & 0.26 / 0.44 \\
\cline{3-7}
& $\mathbf{Average}$ & 0.46 / 1.01 & 0.51 / 1.05 & 0.50 / 1.02 & 0.54 / 1.11 & 0.54 / 1.13 \\
& $\mathbf{Average}$ (TPAD Top-10) & 0.48 / 1.94 & 0.47 / 0.94 & 0.46 / 0.91 & 0.42 / 0.80 & 0.43 / 0.85 \\
\cline{3-7}
& $\mathbf{Worst}$ & 0.68 / 1.58 & 0.73 / 1.48 & 0.72 / 1.48 & 0.80 / 1.62 & 0.80 / 1.67 \\
\bottomrule

\end{tabular}
} 
\label{table6}
\end{table*}
}

\setlength{\tabcolsep}{2.7mm}{
\begin{table*}
\caption{
Performance of stochastic TP models which apply TPAD methods with different Top numbers $\psi$ on HOTEL dataset.
} 
\newcommand{\tabincell}[2]{\begin{tabular}{@{}#1@{}}#2\end{tabular}}
\centering
\resizebox{0.93\linewidth}{!}{ 
\begin{tabular}{@{}ll|ccccc@{}}
\toprule
\multicolumn{2}{l|}{\multirow{2}{*}{\textbf{Stochastic TP Models}}} & \multicolumn{5}{c}{\textbf{Top Number $\psi$}} \\
\cline{3-7}
 & & \textbf{$\psi=5$} & \textbf{$\psi=10$} & \textbf{$\psi=15$} & \textbf{$\psi=20$} & \textbf{$\psi=25$} \\
\toprule

\multirow{2}{*}{STGAT} &  $\mathbf{Best}$ (TPAD Top-$\psi$) & 0.27 / 0.47 & 0.23 / 0.40 & 0.22 / 0.38 & 0.22 / 0.39 & 0.22 / 0.38 \\
& $\mathbf{Average}$ (TPAD Top-$\psi$) & 0.41 / 0.80 & 0.43 / 0.85 & 0.44 / 0.92 & 0.44 / 0.99 & 0.45 / 1.05 \\
\cline{2-7}

\multirow{2}{*}{SGCN} & $\mathbf{Best}$ (TPAD Top-$\psi$) & 0.32 / 0.56 & 0.28 / 0.48 & 0.27 / 0.33 & 0.23 / 0.32 & 0.23 / 0.36 \\
& $\mathbf{Average}$ (TPAD Top-$\psi$) & 0.42 / 0.80 & 0.44 / 0.86 & 0.47 / 0.91 & 0.56 / 1.02 & 0.47 / 0.93 \\
\bottomrule

\end{tabular}
} 
\label{table7}
\end{table*}
}

\begin{figure*}[t]
\includegraphics[scale=0.44]{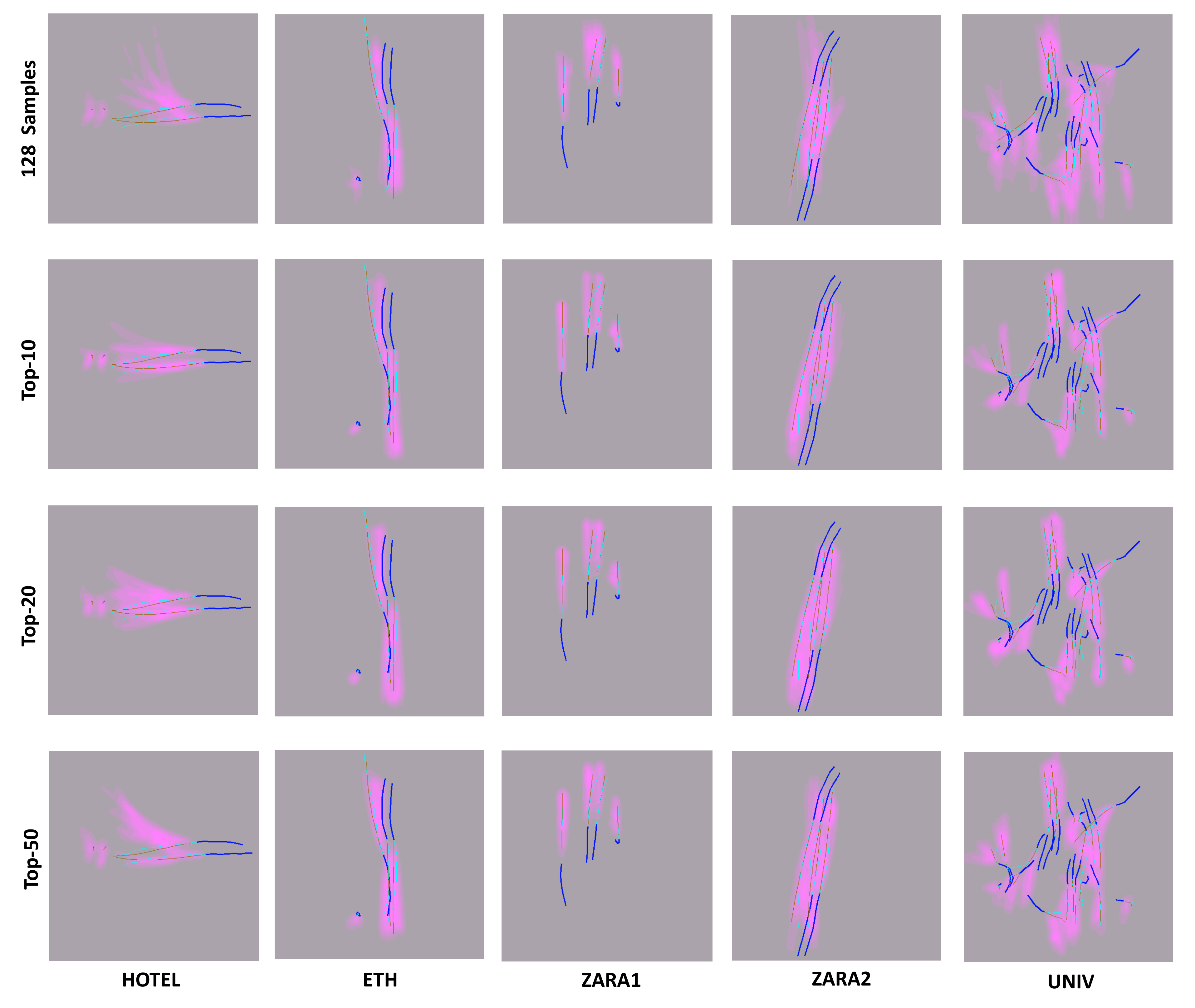}
    \caption{Quality comparison of the diversified prediction results generated by STGAT~\cite{huang2019stgat} and Top-$k$ prediction results filtered by our TPAD method. The observed historical trajectories are shown in dark blue, the ground truth future trajectories are shown in cyan, and the pink areas are covered by 10 trajectory prediction results randomly sampled from all results or the Top-$k$ results.
}
\label{fig:visual-compare}
\end{figure*}

\subsection{TPAD Performance Analysis}\label{section:4.2}
In this part, we examine the ability of TPAD in identifying high-quality trajectory prediction results. We analyze the quality of 50 prediction results provided by the state-of-the-arts stochastic TP models, and compare the performance of the Top-10 prediction results identified by our TPAD method. Table~\ref{table5} gives the $\mathbf{Average}$, $\mathbf{Best}$ and $\mathbf{Worst}$ ADE/FDE score of the stochastic TP models with or without TPAD.

We can observe from Table~\ref{table5} that the quality of the prediction results generated by stochastic TP models varies a lot. The optimal trajectory prediction result assembled by different samples performs very well. However, stochastic TP models can also generate very poor results, much worse than the traditional TP models with deterministic trajectory outputs. In real applications, stochastic TP models are unable to identify high-quality ones from diversified samples, and thus can only randomly select some prediction results for practical tasks. We can see from Table~\ref{table5} that their $\mathbf{Average}$ performance are not good enough (worse than some deterministic TP models). This greatly reduces the practicality of stochastic TP models. 

As for the performance of TPAD method, we notice that TPAD can help stochastic TP models avoid poor prediction results and identify some high-quality ones. Specifically, the $\mathbf{Average}$ performance of the trajectory prediction results filtered by TPAD is much better than that of all results generated by stochastic TP models. Besides, the $\mathbf{Best}$ ADE/FDE score achieved by TPAD is similar to that of the optimal prediction results assembled by all samples. We find that TPAD's $\mathbf{Best}$ ADE/FDE score is generally higher than that of deterministic TP models. This result shows us that the few prediction results filtered by TPAD contain a high-quality solution, and TPAD has the ability to guide stochastic TP models to achieve better practicability. 

Our proposed TPAD method can provide stochastic TP models with effective identification basis, helping them to improve the rationality of the finaly obtained prediction results, which is of great significance in real TP applications.

\subsection{TPAD Parameter Sensitivity Analysis}\label{section:4.3}
In this part, we analyze the effect of two hyperparameters, i.e., the sample number of stochastic TP models and the number of filtered prediction results, on the performance of TPAD method.

\textbf{Varying Sample Number $\Psi$.} We evaluate the performance of TPAD using different sample numbers. We
test 5 different values of $\Psi$, i.e., 20, 40, 60, 80 and 100. For each value of $\Psi$, we report the $\mathbf{Average}$, $\mathbf{Best}$ and $\mathbf{Worst}$ ADE/FDE score of the stochastic TP models with or without TPAD. The results are shown in Table~\ref{table6}. 

As we can see, the $\mathbf{Average}$ and $\mathbf{Worst}$ ADE/FDE scores of stochastic TP models increase, while $\mathbf{Best}$ ADE/FDE scores reduce, with the increasing sample number. This results show us that stochastic TP models can generate not only more accurate but also many worse trajectory prediction results with the increase of $\Psi$. The stochastic TP models can generate diversified prediction results and urgently need an effective identification method to guide them to find high-quality solutions for practical applications.

In addition, we notice that the quality of the prediction results identified by TPAD increases with the increasing sample number. TPAD can effective identify high-quality prediction results from the diversified samples, maintaining a high-performance of the stochastic TP model under different sample numbers.

\textbf{Varying Number of Filtered Prediction Results $\psi$.} We also test the performance of TPAD using 5 different values of $\psi$, i.e., 5, 10, 15, 20 and 25. Results are shown in Table~\ref{table7}. We can see from Table~\ref{table7} that the $\mathbf{Average}$ quality of the prediction results identified by TPAD slightly decreases, and their $\mathbf{Best}$ performance increases, with the increasing $\psi$. This result shows us that TPAD may give high anomaly scores to some high-quality prediction results, but can effective avoid most of worse trajectory prediction results under different values of $\psi$. Our proposed TPAD method may exhibit some errors, but can provide the stochastic TP models with very effective help on the whole.

Figure~\ref{fig:visual-compare} visualizes the trajectory prediction results generated by a stochastic TP model and that filtered by TPAD. Specifically, we compare Top-$10,20,50$ prediction results identified by TPAD with a total of 128 stochastic results generated by STGAT~\cite{huang2019stgat}, and visualize the areas covered by 10 trajectories randomly sampled from them respectively. We can observe from Figure~\ref{fig:visual-compare} that the visualized areas generally cover the ground truth future trajectories, and the Top-$k$ trajectories make the area denser around the ground truth trajectories. This shows that our proposed TPAD method can find a reasonable future trajectory distribution even when $\psi$ is small by leveraging the trajectory AD criterions learned by our proposed trajectory AD model. Also, this results demonstrate the effectiveness of our proposed TPAD.


\section{Conclusion and Future Works}\label{section:5}
In this paper, we introduce the TP, AD and AutoML technique, and design TPAD, an effective TP evaluation method for stochastic TP models. TPAD constructs a trajectory AD model to reasonably analyze the validity of the predicted trajectories. It innovatively applies the trajectory abnormality score to assist users to identify effective prediction results from numbers of candidates. With the help of TPAD, the stochastic TP models can play out their good performance and achieve good application effect. The extensive experimental results demonstrate the effectiveness and significance of TPAD. In this paper, we make the first attempt to utilize the AD technique to solve the identification problems in stochastic TP model, and achieve preliminary success. In future works, we will try to find or design more appropriate AD technique for TP area, to further increase the effectiveness of the TP evaluation methods, so as to assist the stochastic TP model to achieve better practical effect.









\bibliographystyle{cas-model2-names}

\bibliography{cas-refs}

\end{document}